\documentclass[runningheads]{llncs}

\usepackage[T1]{fontenc}
\usepackage{graphicx}
\graphicspath{{./figs/}}
\usepackage{amsmath,amssymb}
\usepackage{booktabs}
\usepackage{array}
\usepackage{xcolor}
\usepackage{cite}
\usepackage{marvosym}
\usepackage[
  colorlinks=true,
  linkcolor=blue,
  citecolor=blue,
  urlcolor=blue
]{hyperref}
\newcolumntype{L}[1]{>{\raggedright\arraybackslash}p{#1}}

\newcommand{\RRPRTotalCandidates}{2000}
\newcommand{\RRPRCompleteCycles}{8}
\newcommand{\RRPRAttemptsPerCycle}{250}
\newcommand{\RRPRStrictSuccesses}{462}
\newcommand{\RRPRStrictSuccessRate}{23.1}
\newcommand{\RRPROtherDataset}{5}

\newcommand{\RRPRCodeErrors}{1348}
\newcommand{\RRPRNoOutput}{185}
\newcommand{\RRPRBestCycleBid}{A6/B190}
\newcommand{\RRPRBestAcc}{0.3676}
\newcommand{\RRPRBestParamsM}{11.8}

\newcommand{\RRPRBestChannels}{[64, 192, 192, 192, 724, 724, 724, 256, 256, 256, 256, 256]}
\newcommand{\RRPREarlyBestAcc}{0.3144}
\newcommand{\RRPRTokenNonPowTwo}{20.2}
\newcommand{\RRPRArchAnyNonPowTwo}{41.8}
\newcommand{\RRPRArchAnyNonPowTwoN}{193}

\newcommand{\RRPRWidthTokenCount}{4961}
\newcommand{\RRPRUniqueWidths}{94}
\newcommand{\RRPRAnyNonPowMean}{0.2144}
\newcommand{\RRPRAllPowMean}{0.2169}
\newcommand{\RRPRAnyNonPowMax}{0.3676}
\newcommand{\RRPRAllPowMax}{0.3276}
\newcommand{\RRPRAZeroBestParamsM}{166.5}
\newcommand{\RRPRAZeroBestAcc}{0.3144}
\newcommand{\RRPRAZeroBestName}{A0/B134}
\newcommand{\RRPRASixBTenTwoParamsM}{5.0}
\newcommand{\RRPRASixBTenTwoAcc}{0.3400}
\newcommand{\RRPRASixBTenTwoName}{A6/B102}
\newcommand{\RRPRErrorRuntime}{426}
\newcommand{\RRPRErrorAPI}{370}
\newcommand{\RRPRErrorMissing}{192}
\newcommand{\RRPRErrorDuplicate}{185}
\newcommand{\RRPRErrorShape}{121}
\newcommand{\RRPRMeanSlope}{9.87\times 10^{-4}}
\newcommand{\RRPRMeanP}{0.043}
\newcommand{\RRPRTopFiveSlope}{4.81\times 10^{-3}}
\newcommand{\RRPRTopFiveP}{0.017}
\newcommand{\RRPRTopTenSlope}{4.02\times 10^{-3}}
\newcommand{\RRPRTopTenP}{0.016}

\newcommand{\RRPRConfoundStrictN}{462}
\newcommand{\RRPRConfoundModalEvalN}{461}

\newcommand{\RRPRConfoundLowresPartialRho}{0.139}
\newcommand{\RRPRConfoundLowresPartialP}{0.003}
\newcommand{\RRPRConfoundLowresPostRho}{0.170}
\newcommand{\RRPRConfoundLowresPostP}{3.8\times 10^{-4}}
\newcommand{\RRPRConfoundMiddlePartialRho}{0.170}
\newcommand{\RRPRConfoundMiddlePartialP}{2.5\times 10^{-4}}

\begin{document}

\title{Scaling Closed-Loop Feature Channel Configuration with LLMs}
\titlerunning{Scaling Channel Configuration}
\author{Tolgay Atinc Uzun\textsuperscript{(\Letter)} \and
Radu Timofte \and
Dmitry Ignatov}
\authorrunning{T. A. Uzun et al.}
\institute{Computer Vision Lab, CAIDAS \& IFI, University of W\"urzburg, Germany\\
\email{t.atincuzun@gmail.com}}
\maketitle

\begin{abstract}
Promising initial results in closed-loop large-language-model-based channel-configuration search demonstrated that neural-network widths can be optimized directly through executable code generation and accuracy feedback. However, those results were obtained from a relatively sparse set of valid evaluations, leaving open whether the observed optimization behavior transfers to a denser sampling regime and whether additional architectural regularities emerge when more generated networks are evaluated. To test this, the same search setting is scaled to \RRPRAttemptsPerCycle{} candidate networks per fine-tuning cycle. The analysis covers \RRPRTotalCandidates{} generated candidates from \RRPRCompleteCycles{} complete cycles, yielding \RRPRStrictSuccesses{} verified CIFAR-100 evaluations after task and metadata filtering. Per-cycle mean accuracy exhibits a positive linear trend with slope $\RRPRMeanSlope$ ($p=\RRPRMeanP$), while the high-performing frontier improves more strongly: the best observed accuracy increases from \RRPREarlyBestAcc{} to \RRPRBestAcc{}, and both the top-5 and top-10 cycle-level means exhibit positive trends. The scaled run also reveals improved parameter efficiency. The best model reaches \RRPRBestAcc{} with \RRPRBestParamsM{}M parameters, compared with an early high-performing model at \RRPRAZeroBestAcc{} with \RRPRAZeroBestParamsM{}M parameters. Beyond accuracy, the larger sample exposes architectural regularities that were difficult to assess from sparse observations. Non-power-of-two channel widths occur in \RRPRArchAnyNonPowTwo\% of verified candidates, and the strongest models share structured channel-allocation patterns characterized by moderate early widths and expanded middle or later blocks. These findings indicate that the channel-search signal observed in the initial study transfers to a larger-data setting and that increased sampling scale turns isolated candidate discovery into a measurable distributional and architectural analysis of LLM-generated channel priors.

\end{abstract}

\section{Introduction}

Neural architecture search (NAS) is usually formulated as optimization over a constrained graph or supernet~\cite{zoph2016neural,liu2018darts}. Channel-number search is a narrower but practically important case: it changes per-layer widths while preserving the executable network skeleton, and it is closely related to efficient-model design and channel-pruning literature~\cite{he2018amc,liu2017slimming,liu2019metapruning,wang2020channelnet}. The preceding study asked whether large language models (LLMs) can perform such search directly over neural-network source code~\cite{uzun2026closedloop}. In that framework, the LLM generates a PyTorch model, a training harness evaluates it, and successful generations are fed back into later fine-tuning cycles.

The preceding result established feasibility on a smaller evaluation sample. Many generated programs may fail, some successful executions may correspond to non-target settings, and the best candidate is an order statistic over a sampled architecture population. Reporting only the final best architecture therefore leaves open the question of whether the search behavior carries over to a larger generation budget.

The present study analyzes a denser run of the same channel-configuration loop. Each complete cycle contains \RRPRAttemptsPerCycle{} generated candidates, compared with a substantially smaller sample in the preceding experiment. The resulting data support three observations: the per-cycle mean accuracy has a positive trend, the high-performing frontier improves under scaling, and later cycles identify substantially more parameter-efficient channel configurations than the early incumbent. The contribution of the present study is therefore a scaled empirical analysis of LLM-driven channel search and the architectural patterns that become visible with more generated candidates.

\noindent\textbf{Contributions.}
First, the study provides a strict candidate-accounting protocol for LLM-generated architecture experiments, distinguishing failed code, missing logs, non-target datasets, and target-valid evaluations. Second, the scaled run recovers and strengthens the accuracy and frontier signals discussed in the preceding study. Third, channel patterns are analyzed at scale using both programmatic extraction and dynamic PyTorch instrumentation of generated networks. Fourth, a self-contained analysis script rebuilds all tables and figures from the raw folders without relying on pre-existing processed data frames.

\section{Related Work}

\noindent\textbf{Neural architecture search and channel configuration.}
Classical NAS formulates architecture design as an optimization problem over a discrete or continuous search space. Reinforcement-learning-based NAS demonstrated that architectures can be learned from validation feedback~\cite{zoph2016neural}, while differentiable methods such as DARTS reduced the cost of search by relaxing architectural decisions into continuous parameters~\cite{liu2018darts}. For efficient convolutional networks, a closely related line of work searches or prunes channel dimensions rather than entire macro-architectures. Network Slimming uses sparsity-induced channel selection~\cite{liu2017slimming}; AMC frames compression as automated policy search~\cite{he2018amc}; MetaPruning predicts weights for pruned channel configurations~\cite{liu2019metapruning}; and AutoSlim searches channel widths in a slimmable network~\cite{wang2020channelnet}. These methods usually operate within predefined search spaces or supernets. The present study instead analyzes channel configurations generated as executable code by a language model, while keeping the scientific focus on channel allocation, parameter efficiency, and proxy accuracy under a scaled candidate budget.

\noindent\textbf{Language models for architecture generation.}
Recent work has explored LLMs as optimizers, code editors, or knowledge sources for architecture design. GPT-style models have been tested for NAS prompting and iterative architecture proposal~\cite{wang2023llmnas,zhang2023gptnas}. EvoPrompting and LLMatic use language-model code generation in evolutionary or quality-diversity search loops~\cite{xu2023evoprompting,nasir2023llmatic}. Other systems use LLMs to transfer design principles~\cite{zhou2024lapt}, perform self-evolution and knowledge-inspired search~\cite{cai2025seki}, or coordinate multiple LLM roles for knowledge-guided search~\cite{li2025collmnas}. A complementary line of work studies the infrastructure needed for generated neural networks to be stored, evaluated, and reused. NNGPT rethinks AutoML around LLM-generated neural-network code under executable API constraints~\cite{kochnev2025nngpt}, while LEMUR and LEMUR2 provide large neural-network datasets and diversity-oriented resources for automated neural design exploration~\cite{lemur2025,lemur2_2026}. The closest prior work to the present study is the closed-loop channel-prior paper, which introduced feedback-driven LLM channel-configuration search and reported promising non-standard channel priors~\cite{uzun2026closedloop}. Compared with these broader architecture-generation and dataset efforts, the present analysis isolates a narrower question: whether the channel-search behavior observed in that prior study persists when the number of generated networks per cycle is increased, and what architectural regularities become measurable from the larger generated population.

\noindent\textbf{Proxy evaluation and short-horizon signals.}
NAS systems often rely on reduced-cost evaluations before committing to expensive full training. Proxy signals may come from short training schedules~\cite{zhou2020econas}, weight sharing through supernetworks~\cite{liu2018darts}, zero-cost indicators computed from a single minibatch~\cite{abdelfattah2021zerocost}, or reduced search spaces. Systematic comparisons show that proxies vary widely in how well they preserve the true ranking of architectures~\cite{yu2020evaluating,zhou2020econas}. Such approximations are useful for ranking many candidates but must be interpreted as search-time evidence rather than final trained performance. The present work follows this logic: one-epoch CIFAR-100 accuracy is treated as a fixed proxy for early learning behavior. The dynamic PyTorch analysis therefore asks which generated channel configurations learn quickly under this proxy, which stage-wise allocations are associated with short-horizon accuracy, and which models improve the accuracy-parameter frontier before full-training confirmation.

\section{Closed-Loop Pipeline and Scaled Evaluation Setting}
\label{sec:pipeline}

The pipeline treats channel search as conditional code generation~\cite{uzun2026closedloop}. The input prompt contains a baseline network, task metadata, hyperparameters, and a target improvement. The LLM then emits a complete neural-network program and training-related blocks. The candidate is evaluated under a fixed one-epoch proxy protocol on CIFAR-100. Valid, evaluated candidates are stored and later used to update the model through parameter-efficient fine-tuning with LoRA~\cite{hu2021lora,dettmers2023qlora}. The experiment uses the OlympicCoder-7B code model~\cite{penedo2025olympiccoder} and an AirNet-like residual convolutional skeleton~\cite{chee2018airnet}. Generated networks and metadata are stored in the LEMUR/NNGPT ecosystem~\cite{lemur2025,lemur2_2026,kochnev2025nngpt}. LoRA hyperparameters follow the standard adapter settings: rank 32, $\alpha=32$, dropout $0.05$, applied to the q, k, v projection modules of the base model; fine-tuning restarts from a fresh copy of the base model at the start of each cycle rather than continuing from the previous adapter.

The study scales the same closed-loop pipeline to \RRPRAttemptsPerCycle{} candidates per cycle, a substantial increase over prior sampling density. The raw experiment directory contains cycles \texttt{A0} through \texttt{A8}. Cycles \texttt{A0}--\texttt{A7} are complete, each with \RRPRAttemptsPerCycle{} candidate folders. Cycle \texttt{A8} is partial (seven CIFAR-100 evaluations) and is excluded from the main analysis.

\noindent\textbf{Compute footprint.}
Across the \RRPRStrictSuccesses{} strict CIFAR-100 evaluations, the per-candidate one-epoch training took a median of $35$~s (IQR $26$--$48$~s) and a total of approximately $7.4$ GPU-hours, with a median peak GPU memory of $21.5$~GB. The 12.8\%--34.4\% strict success rate means roughly two thirds of the per-cycle wall-clock was spent on candidates that did not enter the analysis. This cost is part of why \RRPRCompleteCycles{} cycles of \RRPRAttemptsPerCycle{} candidates each were needed to accumulate the \RRPRStrictSuccesses{} valid CIFAR-100 evaluations, and it frames the search as a sample-efficiency problem as well as an accuracy problem.

\section{Data Accounting Protocol}
\label{sec:accounting}

A target-valid success is defined as a candidate in \texttt{A0}--\texttt{A7} with a valid \texttt{\{epoch\_number\}.json} file (under the one-epoch protocol, \texttt{1.json}) and an \texttt{eval\_info.json} confirming evaluation on \texttt{cifar-100}. The dataset check removes false-positive successful evaluations on other tasks. This rule yields \RRPRStrictSuccesses{} strict CIFAR-100 evaluations out of \RRPRTotalCandidates{} generated candidates.

\begin{figure}[t]
\centering
\includegraphics[width=0.86\textwidth]{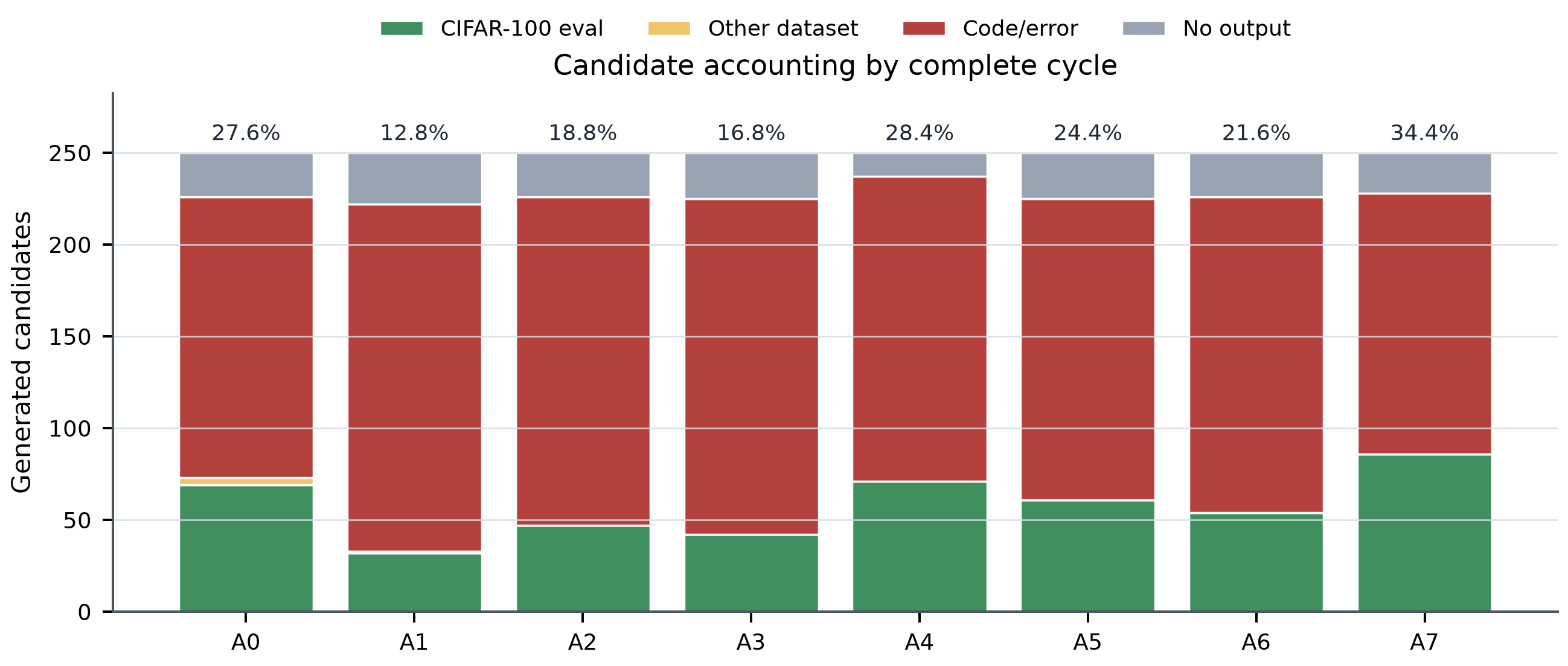}
\caption{Candidate accounting over the \RRPRCompleteCycles{} complete cycles. The figure separates target-valid CIFAR-100 evaluations from non-target evaluations, code errors, and missing outputs.}
\label{fig:accounting}
\end{figure}

Table~\ref{tab:accounting} and Fig.~\ref{fig:accounting} summarize the complete accounting. Across complete cycles, \RRPRCodeErrors{} candidates terminate with an error trace, \RRPRNoOutput{} have no logged output, and \RRPROtherDataset{} are successful evaluations on a dataset other than CIFAR-100. The non-target evaluations are not failures of code execution, but they are excluded from the CIFAR-100 analysis because they do not answer the target channel-search question.

\begin{table}[t]
\centering
\caption{Per-cycle candidate accounting for the complete cycles. Each cycle contains \RRPRAttemptsPerCycle{} generated candidates. Only the strict CIFAR-100 column is used for accuracy and architecture claims.}
\label{tab:accounting}
\setlength{\tabcolsep}{5pt}
\small
\begin{tabular}{lrrrrr}
\toprule
Cycle & CIFAR-100 & Other & Code/err & No output & Rate (\%)\\
\midrule
A0 & 69 & 4 & 153 & 24 & 27.6\\
A1 & 32 & 1 & 189 & 28 & 12.8\\
A2 & 47 & 0 & 179 & 24 & 18.8\\
A3 & 42 & 0 & 183 & 25 & 16.8\\
A4 & 71 & 0 & 166 & 13 & 28.4\\
A5 & 61 & 0 & 164 & 25 & 24.4\\
A6 & 54 & 0 & 172 & 24 & 21.6\\
A7 & 86 & 0 & 142 & 22 & 34.4\\
\bottomrule
\end{tabular}
\end{table}

Overall, \RRPRStrictSuccessRate\% of generated candidates become target-valid CIFAR-100 evaluations under this filtering rule. The unsuccessful candidates are used for error analysis, while the successfully trained CIFAR-100 models are used for the subsequent accuracy, efficiency, channel-configuration, and dynamic PyTorch analyses.

\section{Accuracy Trends Under Scaling}

Table~\ref{tab:accuracy} summarizes the strict CIFAR-100 results. The means lie in a narrow band from 0.2107 to 0.2203. A linear fit to the per-cycle means gives a positive slope of $\RRPRMeanSlope$ accuracy points per cycle ($p=\RRPRMeanP$), indicating an upward trend at the cycle-mean level across eight cycle-level observations. Individual candidates are highly variable, so the stronger empirical signal is found in the upper tail of the distribution rather than in a per-candidate regression.

\begin{table}[t]
\centering
\caption{Accuracy statistics for strict CIFAR-100 candidates. Top-5 mean is computed within each cycle and highlights the high-performing tail.}
\label{tab:accuracy}
\begin{tabular}{lrrrrrr}
\toprule
Cycle & N & Mean & Median & Q90 & Max & Top-5 mean\\
\midrule
A0 & 69 & 0.2122 & 0.2148 & 0.2616 & 0.3144 & 0.2852\\
A1 & 32 & 0.2134 & 0.2175 & 0.2676 & 0.2953 & 0.2788\\
A2 & 47 & 0.2107 & 0.2109 & 0.2568 & 0.2992 & 0.2747\\
A3 & 42 & 0.2166 & 0.2115 & 0.2745 & 0.3061 & 0.2819\\
A4 & 71 & 0.2180 & 0.2246 & 0.2578 & 0.3276 & 0.2971\\
A5 & 61 & 0.2187 & 0.2165 & 0.2646 & 0.3495 & 0.3103\\
A6 & 54 & 0.2203 & 0.2193 & 0.2708 & 0.3676 & 0.3133\\
A7 & 86 & 0.2156 & 0.2138 & 0.2652 & 0.3068 & 0.3008\\
\bottomrule
\end{tabular}
\end{table}

\begin{figure}[t]
\centering
\includegraphics[width=0.86\textwidth]{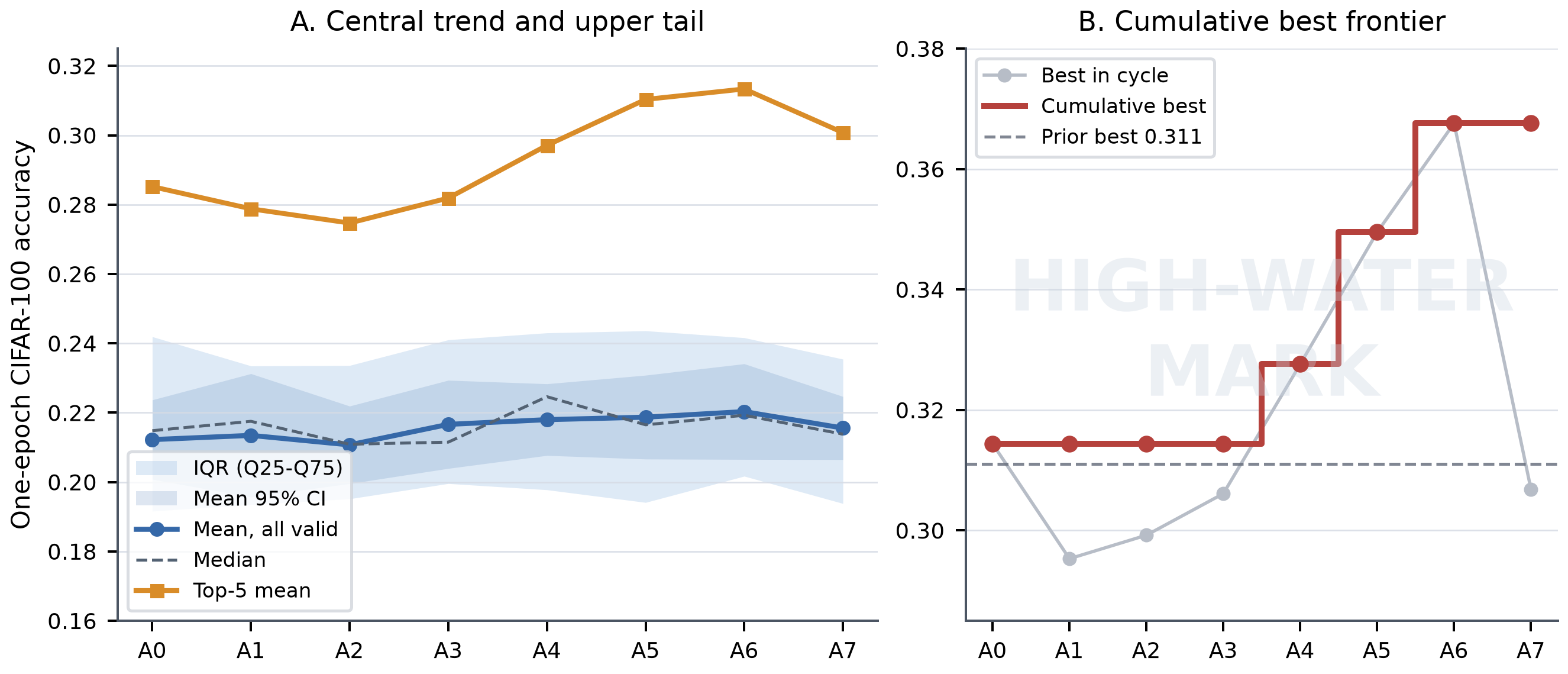}
\caption{Accuracy trends and frontier dynamics. Left: per-cycle mean accuracy for all strict CIFAR-100 candidates with a 95\% confidence band, interquartile shading, median line, and top-5 mean. Right: the per-cycle best and cumulative high-water mark; the dashed reference is the best accuracy reported in the preceding study~\cite{uzun2026closedloop}.}
\label{fig:frontier}
\end{figure}

The primary scaling signal is in the frontier. The best strict CIFAR-100 model in \texttt{A0} reaches \RRPREarlyBestAcc{}. Later cycles improve the high-water mark to \RRPRBestAcc{} at \RRPRBestCycleBid{}. Fig.~\ref{fig:frontier} shows this distinction: the central mass of the distribution remains broad and noisy, while the upper tail improves more clearly. Linear fits to the top-5 and top-10 per-cycle means give slopes of $\RRPRTopFiveSlope$ ($p=\RRPRTopFiveP$) and $\RRPRTopTenSlope$ ($p=\RRPRTopTenP$), respectively. This supports the interpretation that scaling the candidate budget primarily reveals improved high-performing candidates rather than a uniform shift of the entire generated population.

These results show that the search improves the high-performing frontier over successive cycles. The best model is an order statistic that depends on sampling budget, and the scaled experiment makes that dependence visible: later samples include rare high-performing configurations that were less accessible from a smaller sample.

The trend p-values are computed on eight cycle-level points; bootstrap 95\% confidence intervals on the per-cycle means and maxima are wide and overlap across cycles. The most stable statistical evidence is the positive linear trend of the top-5 and top-10 per-cycle means ($p<0.02$), rather than the per-cycle mean alone.

\section{Efficiency of the Discovered Frontier}

The scaled run also shows a parameter-efficiency effect. \RRPRAZeroBestName{} reaches the highest A0 accuracy (\RRPRAZeroBestAcc{}) but is the largest A0 model at \RRPRAZeroBestParamsM{}M parameters --- the early cycle's best raw score came at a high parameter cost (the A0 median is 6.5~M, the 75th percentile 19~M). Late cycles recover the same or higher accuracy at much smaller scales: \RRPRASixBTenTwoName{} reaches \RRPRASixBTenTwoAcc{} with \RRPRASixBTenTwoParamsM{}M parameters, and the global best \RRPRBestCycleBid{} reaches \RRPRBestAcc{} with \RRPRBestParamsM{}M parameters. The shift is also visible at the population level: 26 of 69 A0 strict candidates are Pareto-dominated by \RRPRBestCycleBid{} (lower accuracy at higher parameter counts), and the per-cycle maximum at matched parameter budgets is consistently higher in cycles A4--A7 than in A0--A3. Thus the later frontier does not merely raise raw accuracy; it identifies channel allocations that dominate the early frontier across the full parameter range.

\begin{figure}[t]
\centering
\includegraphics[width=0.86\textwidth]{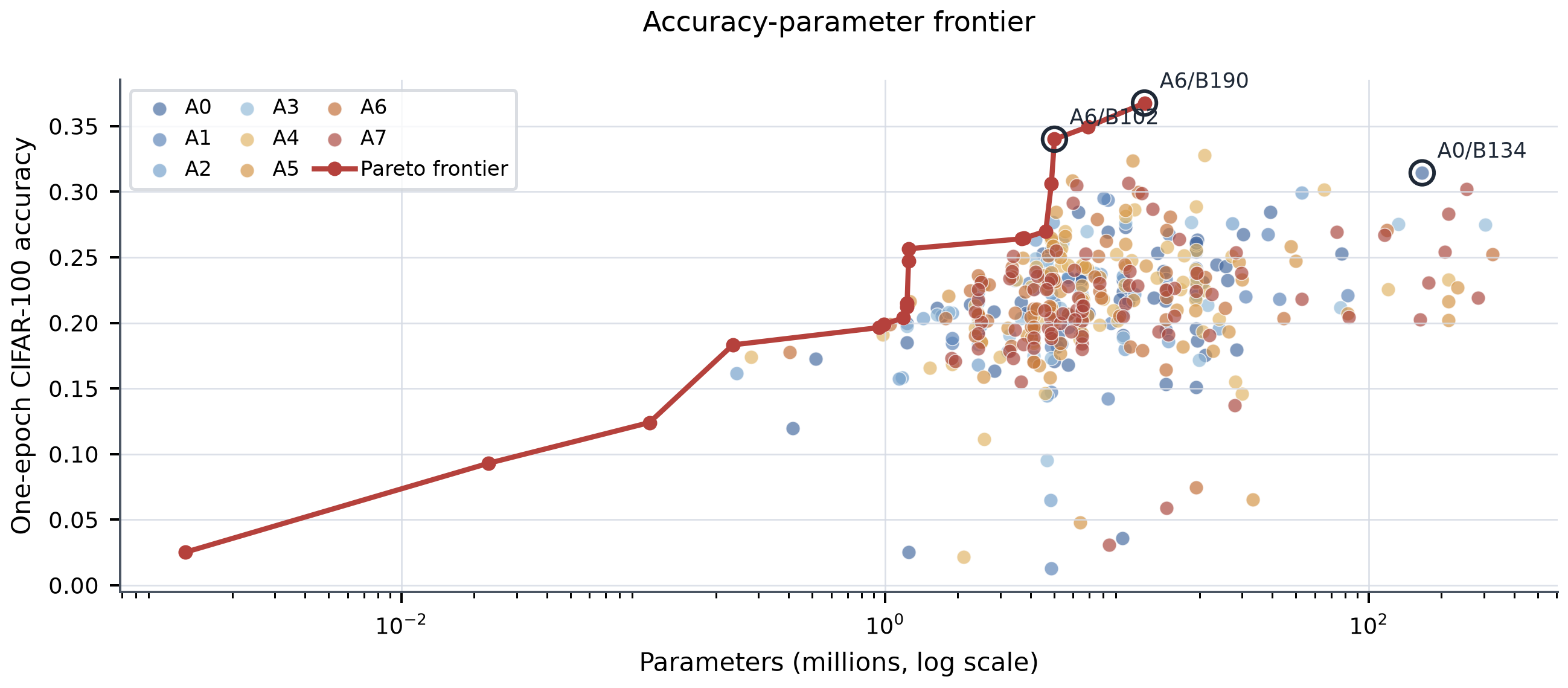}
\caption{Accuracy versus parameter count. Points are strict CIFAR-100 candidates, colored by generation cycle. The red curve is the empirical Pareto frontier. Late cycles include candidates that improve over the early high-parameter incumbent while using far fewer parameters.}
\label{fig:efficiency}
\end{figure}

Fig.~\ref{fig:efficiency} shows the empirical Pareto frontier. This analysis supports comparison beyond the single best accuracy value. A scaled search should be evaluated not only by whether it finds a numerically similar maximum, but also by whether it recovers comparable trade-offs between accuracy and model size.

\section{Channel Configuration Patterns}

The analysis script extracts convolutional output widths by executing each generated \texttt{new\_nn.py} under a lightweight \texttt{torch.nn}-compatible stub module, recording constructor arguments for \texttt{Conv2d}, \texttt{BatchNorm2d}, and \texttt{Linear} without GPU dependencies. Extraction succeeds for all \RRPRStrictSuccesses{} strict CIFAR-100 candidates, yielding \RRPRWidthTokenCount{} convolutional width tokens with \RRPRUniqueWidths{} distinct values.

\subsection{Non-standard Widths}

Non-power-of-two widths appear in \RRPRArchAnyNonPowTwoN{} of \RRPRStrictSuccesses{} strict architectures, or \RRPRArchAnyNonPowTwo\% at the architecture level; at the token level, \RRPRTokenNonPowTwo\% of convolutional widths are non-power-of-two. Power-of-two values remain dominant --- the most frequent widths are conventional values such as 256, 512, 64, 128, and 192. Architectures with at least one non-power-of-two width have mean accuracy \RRPRAnyNonPowMean{}, compared with \RRPRAllPowMean{} for all-power-of-two architectures, a negligible mean difference; the distinction is visible at the frontier, where the best all-power-of-two architecture reaches \RRPRAllPowMax{} and the best architecture with at least one non-power-of-two width reaches \RRPRAnyNonPowMax{}.

\subsection{Width-Position Patterns}

\begin{figure}[t]
\centering
\includegraphics[width=0.86\textwidth]{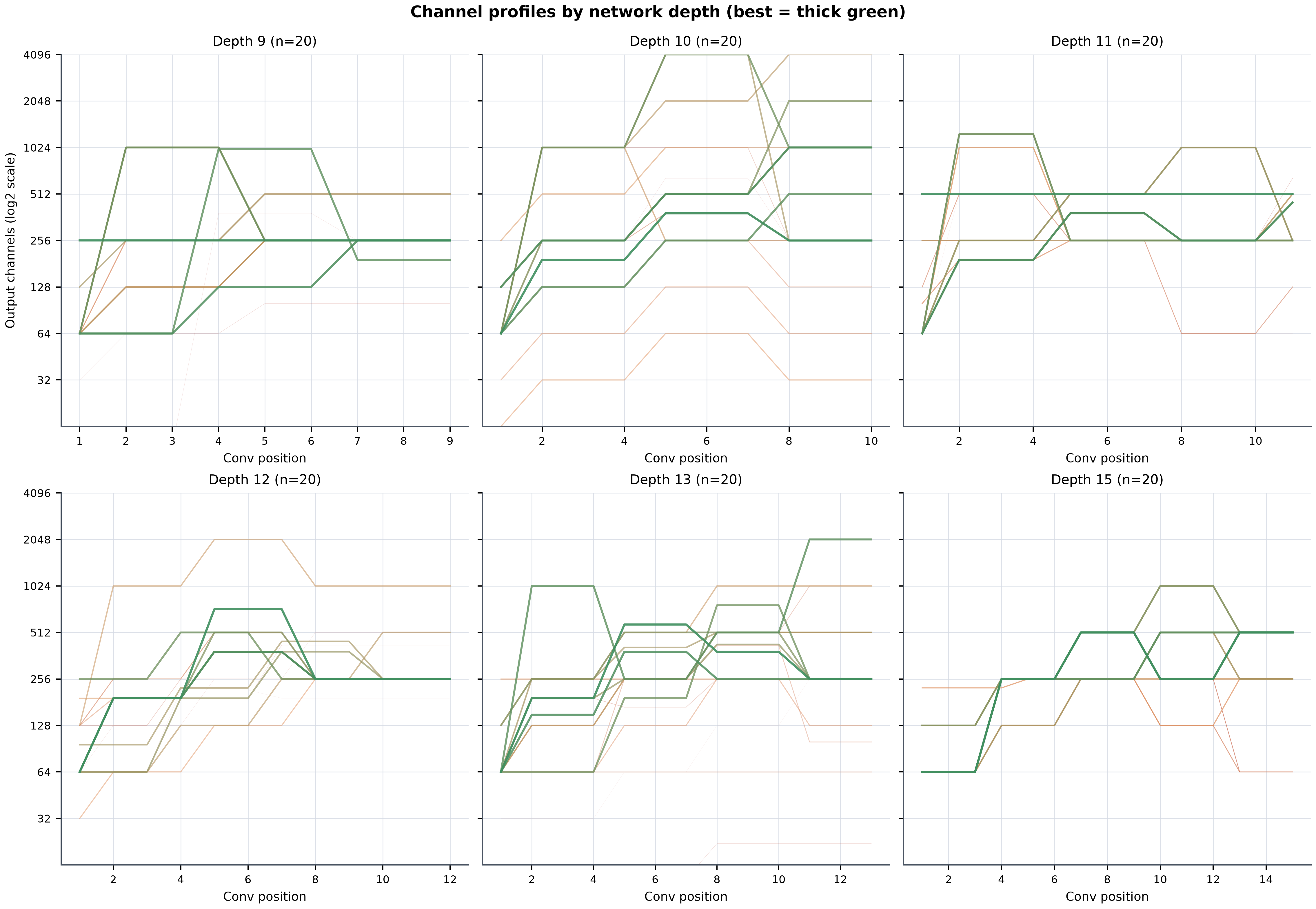}
\caption{Channel profiles faceted by network depth. Within each depth group, lines are colored by accuracy (thin red = low, thick green = high). The pattern of wider middle blocks for better models holds across depth values.}
\label{fig:depthconfound}
\end{figure}

\begin{figure}[t]
\centering
\includegraphics[width=0.86\textwidth]{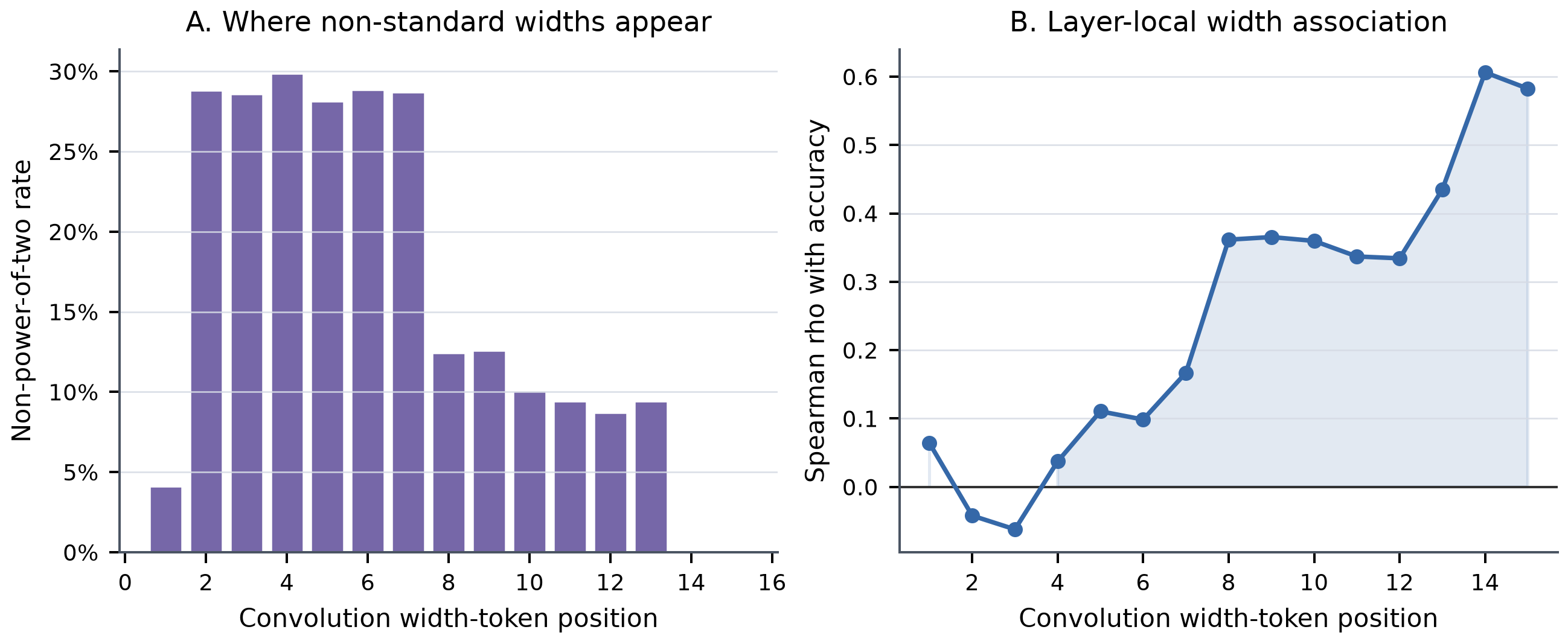}
\caption{Layer-local channel statistics. Left: non-power-of-two width rate by convolutional layer index. Right: Spearman correlation between layer width and validation accuracy. Later layers show stronger positive association with accuracy than the input stem.}
\label{fig:layerstats}
\end{figure}

Table~\ref{tab:top} lists the top-performing candidates. The best model (\RRPRBestCycleBid{}, 0.3676) has channel sequence \RRPRBestChannels{} --- a 64-channel stem, three 192-channel layers, three 724-channel layers, and five 256-channel layers in run-length form. The pattern is blockwise: moderate early capacity, a wider middle section, and a controlled later representation. Fig.~\ref{fig:depthconfound} shows that this pattern --- better trajectories tend toward wider middle blocks --- holds across network depths, and Fig.~\ref{fig:layerstats} shows the corresponding layer-local statistics. Non-power-of-two widths concentrate in early-middle positions, especially layers 2--7, and the association between width and accuracy is weaker near the stem and stronger in later positions. The per-layer tests are not independent, but the result shows that the scaled run produces structured, non-random width variation.

\subsection{Dynamic PyTorch Evidence for Capacity Placement}

The generated code permits a direct dynamic analysis with real PyTorch. Each strict CIFAR-100 network is instantiated, evaluated on a dummy CIFAR-shaped tensor, and instrumented with hooks on convolutional, linear, normalization, and pooling layers. This yields observed tensor shapes, parameter counts, and approximate multiply-accumulate counts (MACs) for all \RRPRStrictSuccesses{} verified networks. These measurements are architectural properties of the generated programs, while their association with accuracy should be interpreted under the fixed one-epoch proxy protocol.

Two complementary splits of the convolutional layers are used below. \textbf{Stage-wise} labels a conv layer as \emph{early}, \emph{middle}, or \emph{late} by its position (first, middle, and last third of layers, respectively). \textbf{Resolution-wise} labels a layer by the spatial resolution of its output feature map (e.g., 32$\times$32, 16$\times$16, 8$\times$8). The distinction matters because a layer's \emph{width} (number of output channels) and its \emph{feature map} (the spatial tensor (channels $\times$ height $\times$ width)) are different objects --- a 192-channel layer at 32$\times$32 is far more expensive than the same 192 channels at 8$\times$8, because the convolution is applied at every spatial position. Resolution-wise grouping is therefore the more informative split for capacity analysis.

\begin{figure}[t]
\centering
\vspace{-3mm}
\includegraphics[width=0.86\textwidth]{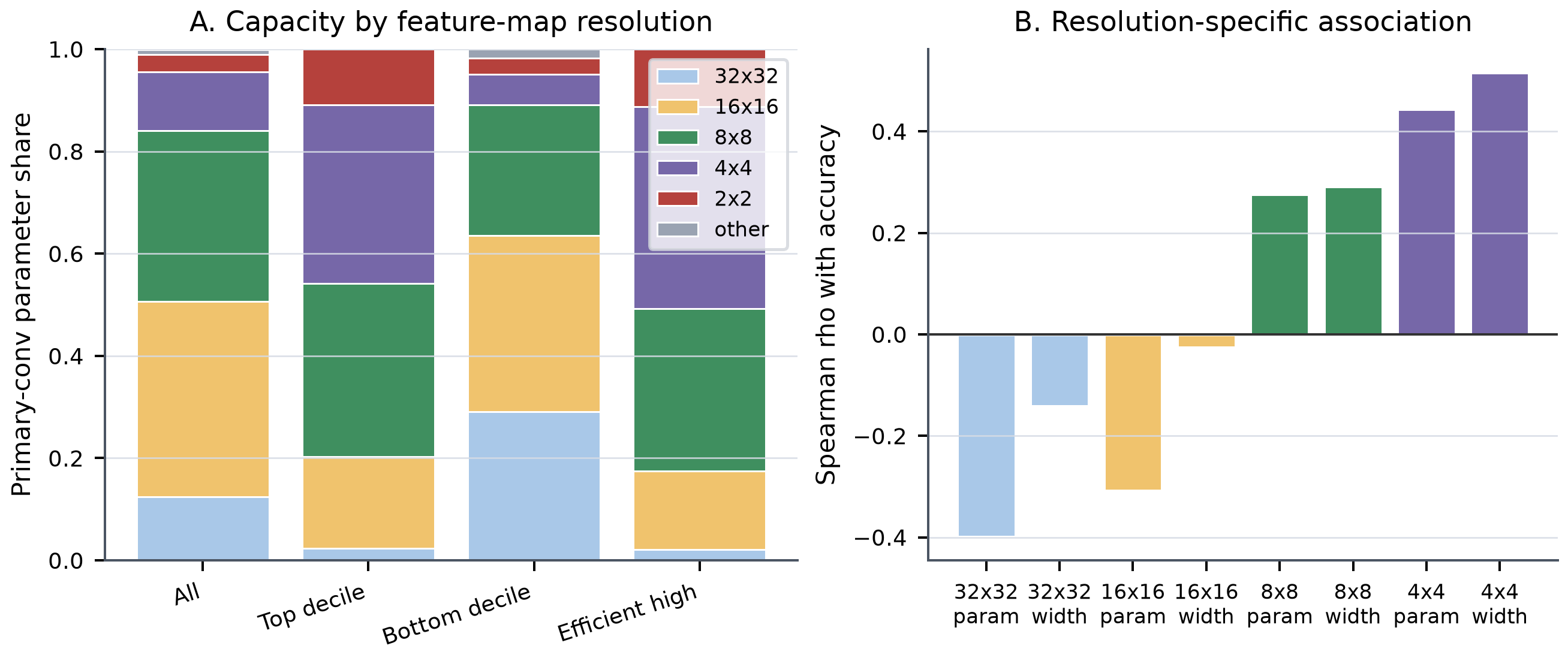}
\vspace{-2mm}
\caption{Resolution-aware dynamic analysis. Left: primary-convolution parameter allocation by output feature-map resolution. Top-decile and efficient high-accuracy candidates allocate more capacity to lower-resolution stages than bottom-decile candidates. Right: resolution-specific Spearman associations with one-epoch accuracy.}
\label{fig:resolutionaware}
\vspace{-3mm}
\end{figure}

The resolution-wise analysis is shown in Fig.~\ref{fig:resolutionaware}. Top-decile proxy models allocate substantially less primary-convolution parameter share to \texttt{32x32} layers and substantially more to lower-resolution layers. In the full verified set (N=\RRPRStrictSuccesses{}), the share of primary-convolution parameters located at \texttt{8x8} or lower resolution has Spearman correlation $\rho=0.477$ with one-epoch accuracy ($p<10^{-26}$). Among the N=119 models that contain \texttt{4x4} convolutions, the mean \texttt{4x4} channel width has a similar-magnitude association ($\rho=0.516$, $p<10^{-8}$); the larger coefficient reflects the more restricted parameter range of this subset, not a stronger underlying signal.

\begin{figure}[t]
\centering
\vspace{-3mm}
\includegraphics[width=0.86\textwidth]{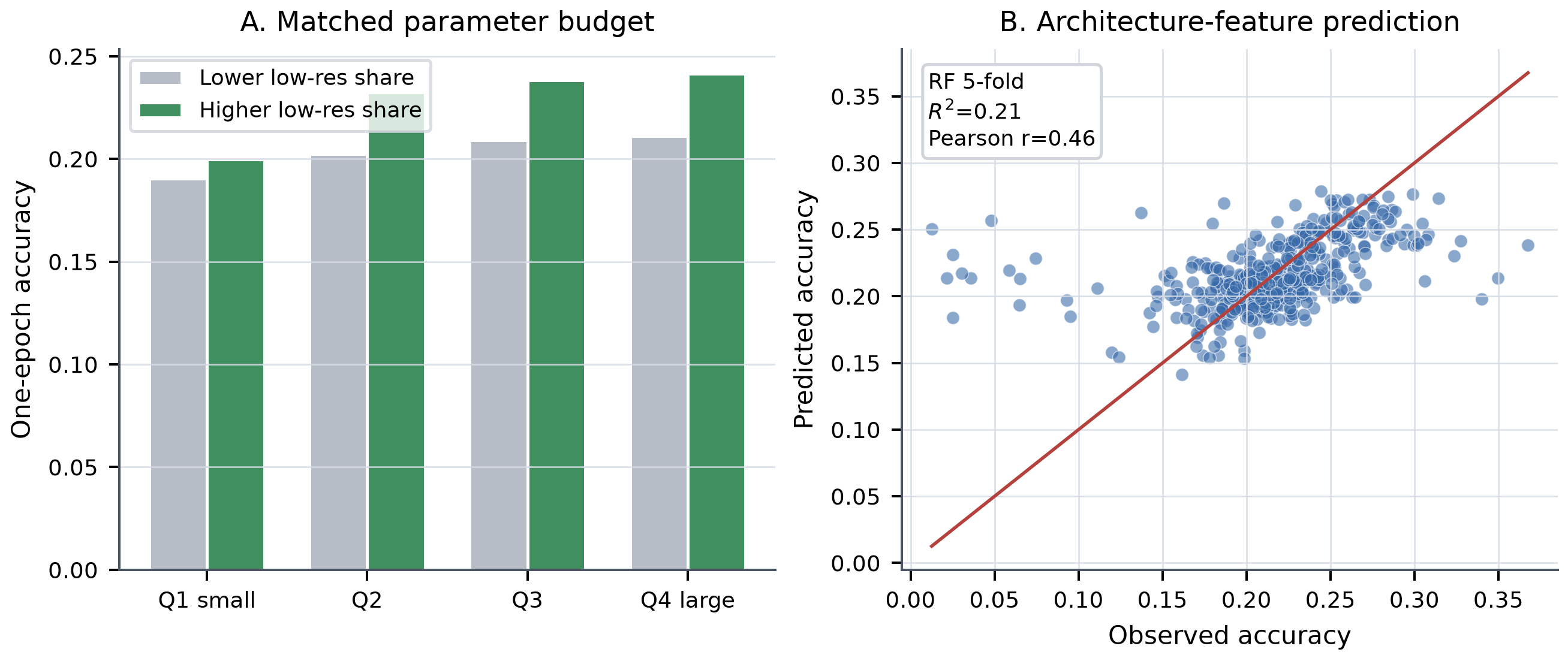}
\vspace{-2mm}
\caption{Matched-budget and predictive analyses. \textbf{A}: within each parameter-count quartile, candidates with higher low-resolution primary-convolution parameter share (green) outperform the lower half (gray), so the resolution effect is not a size artifact. \textbf{B}: predicted vs.\ observed one-epoch accuracy from a leave-cycle-out random forest on architecture features ($R^2=0.22$, Pearson $r=0.48$).}
\label{fig:budgetpredictive}
\vspace{-3mm}
\end{figure}

To test whether the resolution effect is a proxy for larger models, candidates are compared within matched parameter-count quartiles. Quartile here means the \RRPRStrictSuccesses{} strict candidates are sorted by parameter count and split into four equal-sized groups Q1--Q4 (Q1 = smallest 25\%, Q4 = largest 25\%); within each quartile, the candidates are then split into a \emph{lower} and a \emph{higher} half by their low-resolution primary-convolution parameter share. Fig.~\ref{fig:budgetpredictive}~A shows the result: the higher low-resolution group wins in all four quartiles, with a gap of approximately 0.030 accuracy points in Q2, Q3, and Q4 and one-sided Welch tests giving $p<10^{-3}$ in those three bins. If the resolution effect were only a size artifact, the two bars within each quartile would be roughly equal; they are not, so the resolution effect carries information beyond total parameter count.

A complementary question is whether architecture features extracted from generated code can \emph{predict} one-epoch proxy performance at all. A random-forest regressor is trained on dynamic architecture features (resolution shares, layer-position ratios, log-parameter count, depth, and so on) and asked to predict each candidate's one-epoch accuracy, with leave-cycle-out cross-validation so that the model is tested only on cycles it has not seen. Fig.~\ref{fig:budgetpredictive}~B shows the result: each dot is one candidate, the $x$-coordinate is the observed accuracy and the $y$-coordinate is the predicted accuracy, with the red diagonal marking perfect prediction. The points follow the diagonal loosely: leave-cycle-out $R^2=0.219$ with Pearson $r=0.476$ and Spearman $\rho=0.592$ (RMSE $0.040$, roughly 4 accuracy points). The most important features are low-resolution parameter share, log-parameter count, middle-over-early MAC allocation, late-stage mean width, and depth. The remaining $\sim$78\% of the variance is consistent with the noise floor of one-epoch CIFAR-100 training and LLM sampling: the signal is informative about short-horizon learning without fully determining the proxy outcome.

\subsection{Confound Audit for Training and Hidden Code Changes}

The performance signal could in principle come from three places: (i) the channel configuration; (ii) drift in the training protocol, evaluation metadata, or non-channel architectural choices (stride, downsampling, pooling, dropout, weight decay, Nesterov); or (iii) the one-epoch training stochasticity itself. This section audits (ii) and shows that the channel signal in (i) survives the controls.

\begin{figure}[t]
\centering
\includegraphics[width=0.86\textwidth]{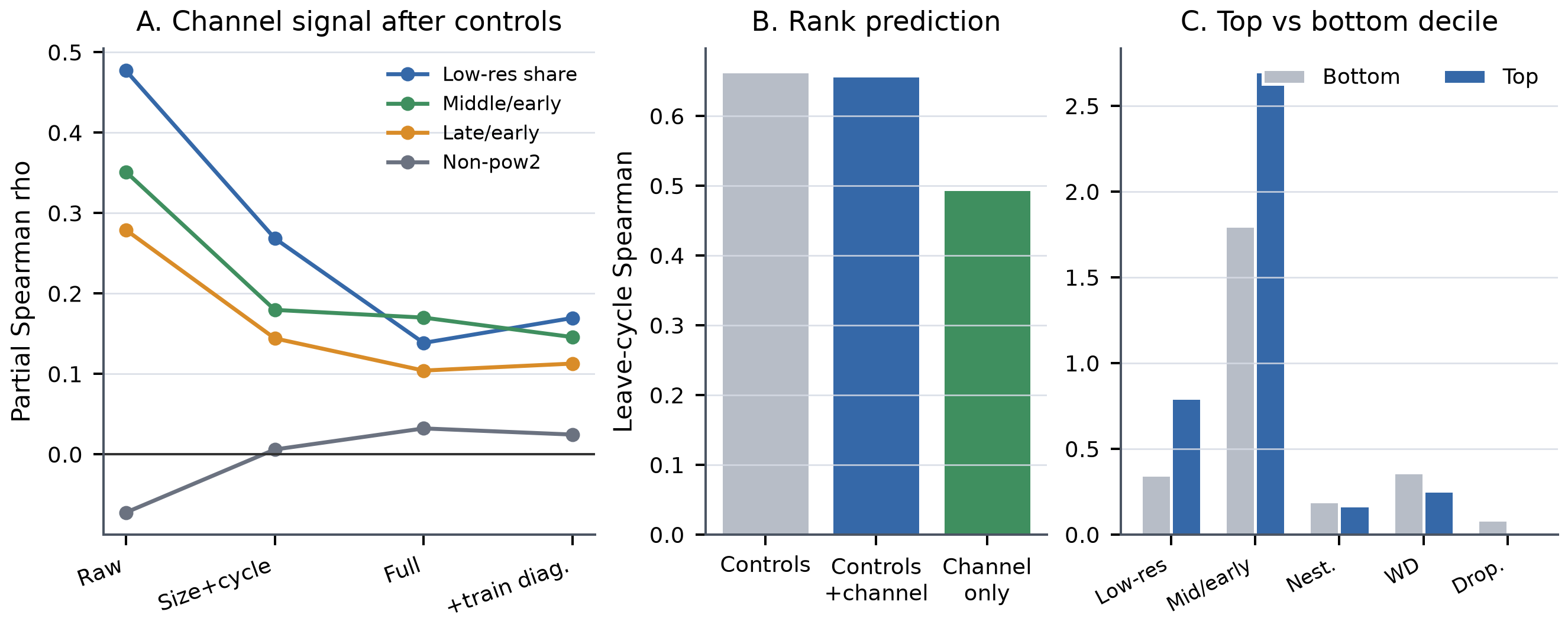}
\caption{Confound audit for the channel-configuration claims. \textbf{A}: partial Spearman $\rho$ of four channel features under progressively stronger control sets. \textbf{B}: leave-cycle-out rank prediction for non-channel controls, controls plus channel features, and channel features only. \textbf{C}: top vs.\ bottom accuracy decile for two channel features and three hidden-code flags. Panel details are described in the text.}
\label{fig:confoundaudit}
\end{figure}

The audit is summarized in Fig.~\ref{fig:confoundaudit}. \textbf{Panel A} plots the partial Spearman $\rho$ between each of four channel features and one-epoch accuracy under progressively stronger controls --- \emph{Raw}, \emph{Size+cycle}, \emph{Full} (size, cycle, depth, MACs, layer-count features, stride, downsampling, pooling, evaluation settings, and hidden-code flags), and \emph{+train diag.} (full controls plus train loss and gradient norm). The blue line for low-resolution share starts at $\rho\approx 0.48$ under \emph{Raw} and only drops to $\rho\approx 0.17$ under \emph{+train diag.}; the other channel lines follow a similar pattern, so the signal does not collapse as controls tighten. \textbf{Panel B} compares three feature sets for leave-cycle-out rank prediction: non-channel \emph{Controls}, \emph{Controls + channel}, and \emph{Channel only}. The \emph{Channel only} bar (green) is taller than the \emph{Controls} bar (gray), and adding controls on top of channels does not improve further. \textbf{Panel C} splits the \RRPRStrictSuccesses{} strict candidates by accuracy into \emph{deciles} (10\% groups, $\sim$46 candidates each) and compares the bottom-decile mean (gray) to the top-decile mean (blue) for two channel features (\emph{Low-res}, \emph{Mid/early}) and three hidden-code flags (\emph{Nest.} = Nesterov, \emph{WD} = weight decay, \emph{Drop.} = dropout). The channel features differ sharply between top and bottom; the hidden-code features do not. The accuracy gap is in the channels, not in the training code.

Under the controls described in Panel A, primary low-resolution parameter share remains positively associated with one-epoch accuracy ($\rho=\RRPRConfoundLowresPartialRho$, $p=\RRPRConfoundLowresPartialP$), and the middle-over-early channel-allocation ratio remains similarly positive ($\rho=\RRPRConfoundMiddlePartialRho$, $p=\RRPRConfoundMiddlePartialP$). Adding post-training diagnostics such as train loss and gradient norm leaves the low-resolution association positive ($\rho=\RRPRConfoundLowresPostRho$, $p=\RRPRConfoundLowresPostP$); these diagnostics are partly a consequence of the architecture and serve as a sensitivity check rather than as primary controls.

\begin{table}[t]
\centering
\caption{Top strict CIFAR-100 candidates. Channel configurations are shown in run-length form.}
\label{tab:top}
\begin{tabular}{lrrrL{6.0cm}}
\toprule
Candidate & Accuracy & Params & Depth & Channel configuration\\
\midrule
A6/B190 & 0.3676 & 11.8M & 12 & 64, 192x3, 724x3, 256x5\\
A5/B99 & 0.3495 & 6.9M & 12 & 64, 192x3, 384x3, 256x5\\
A6/B102 & 0.3400 & 5.0M & 10 & 64, 192x3, 384x3, 256x3\\
A4/B152 & 0.3276 & 20.9M & 10 & 128, 256x3, 512x3, 1024x3\\
A5/B146 & 0.3235 & 10.6M & 13 & 64, 192x3, 576x3, 384x3, 256x3\\
A0/B134 & 0.3144 & 166.5M & 16 & 256, 512x3, 1024x3, 2048x3, 1024x3, 2048x3\\
A5/B145 & 0.3084 & 6.0M & 13 & 64, 150x3, 384x3, 256x6\\
A7/B43 & 0.3068 & 10.1M & 15 & 64x3, 256x3, 512x3, 256x3, 512x3\\
\bottomrule
\end{tabular}
\end{table}

\section{Execution Validity and Failure Modes}

Failure modes are part of the generated-code search process. Among the \RRPRCodeErrors{} code/error candidates, the most frequent categories are generic runtime errors (\RRPRErrorRuntime{}), API signature mismatches (\RRPRErrorAPI{}), missing required functions (\RRPRErrorMissing{}), checksum duplicates (\RRPRErrorDuplicate{}), and shape mismatches (\RRPRErrorShape{}); these labels are deterministic regex categories over error traces and are descriptive rather than a complete compiler taxonomy. Scaled studies of LLM-based NAS should preserve raw generated code, training logs, evaluation metadata, rejected error traces, and enough identifiers to reconstruct candidate lineage. In this run, strict filtering removed non-target dataset evaluations and \texttt{1.json}-only records that would otherwise distort the accuracy trajectory.

\section{Reproducibility Statement}
\label{sec:reproducibility}

Every quantitative claim in this paper is regenerated from the raw experiment directory by a single self-contained analysis script. The script parses the candidate folders of cycles \texttt{A0}--\texttt{A8}, applies the strict filtering rule of Sect.~\ref{sec:accounting} (\texttt{1.json} plus \texttt{eval\_info.json} verification of \texttt{cifar-100}), and rebuilds all tables, all figures, and the numeric macros injected into this document through \texttt{generated/rrpr\_numbers.tex} and \texttt{generated/confound\_numbers.tex}, without relying on any pre-existing processed data frames. The generation stack is fixed and stated in Sect.~\ref{sec:pipeline}: OlympicCoder-7B~\cite{penedo2025olympiccoder} as the base code model; LoRA adapters with rank 32, $\alpha=32$, dropout $0.05$ on the q, k, v projection modules; a fresh copy of the base model at the start of each cycle; and a fixed one-epoch CIFAR-100 proxy evaluation for every candidate. Generated networks, evaluation metadata, and candidate lineage identifiers are stored in the LEMUR/NNGPT ecosystem~\cite{lemur2025,lemur2_2026,kochnev2025nngpt}. The preserved artifact comprises the raw candidate folders (generated \texttt{new\_nn.py} programs, training logs, error traces, \texttt{1.json} and \texttt{eval\_info.json} records for all \RRPRTotalCandidates{} candidates, including failures and non-target evaluations), the analysis script, and the generated macro files, so that the complete accounting of Table~\ref{tab:accounting} --- not only the successes --- can be independently re-derived. The compute footprint of the strict evaluations is reported in Sect.~\ref{sec:pipeline} (approximately $7.4$ GPU-hours in total; median $35$~s per one-epoch candidate; median peak GPU memory $21.5$~GB).

\section{Limitations and Future Work}

\noindent\textbf{Scope of the experiment.}
The experiment is one execution of the closed loop with OlympicCoder-7B, CIFAR-100, and a one-epoch proxy. Other code LLMs may yield different channel priors, and other datasets or longer schedules may re-rank candidates. The confound audit controls for non-channel code variation, but not for LLM identity, dataset identity, or training horizon. The generated samples are also not i.i.d., since later cycles depend on LoRA fine-tuning from earlier cycles.

\noindent\textbf{Sampling-budget considerations.}
A0's best (0.3144) and A6's best (0.3676) differ by 0.053 absolute accuracy. With the observed A0 mean of 0.212 and per-cycle standard deviation of about 0.030, the expected maximum of a 250-sample distribution is roughly 0.31, while the expected maximum of the 54-sample A6 cycle is about 0.27. A second run with the same prompt, LLM, and protocol would produce a different frontier and would help separate a true accuracy shift from chance on a single realization.

\noindent\textbf{Partial signal from architecture features.}
Resolution-specific Spearman values (0.48--0.52) and the leave-cycle-out random-forest model ($R^2=0.219$, Spearman $\rho=0.592$) capture an informative but partial signal: roughly three quarters of one-epoch accuracy variance is not explained by the extracted features, consistent with one-epoch training stochasticity and LLM sampling being substantial noise sources. Width tokens are extracted from constructor arguments under a \texttt{torch.nn}-compatible stub, and tensor shapes and MACs are measured on a single dummy forward pass, so they approximate rather than replicate the trained network's behavior. No candidate was trained for more than one epoch, and a full-training run on the top candidates would clarify whether one-epoch rankings correlate with final rankings.

\noindent\textbf{Future work.}
Future work should test whether one-epoch rankings correlate with final rankings, and should perform controlled channel-allocation interventions in which depth and parameter budget are held fixed while channels are moved from early to later stages or between power-of-two and non-power-of-two middle widths. The partial \texttt{A8} folder also indicates that robust experiment orchestration remains an open practical concern.

\section{Conclusion}

The scaled experiment shows that closed-loop LLM channel search improves both accuracy and parameter efficiency with increased candidate budget. From \RRPRTotalCandidates{} generated candidates in \RRPRCompleteCycles{} complete cycles, \RRPRStrictSuccesses{} are strict CIFAR-100 evaluations; the high-performing frontier improves from \RRPREarlyBestAcc{} to \RRPRBestAcc{}, with \RRPRASixBTenTwoName{} reaching \RRPRASixBTenTwoAcc{} at \RRPRASixBTenTwoParamsM{}M parameters and \RRPRAZeroBestName{} at \RRPRAZeroBestParamsM{}M. The confound audit confirms that the main training protocol is effectively fixed for \RRPRConfoundModalEvalN{} of \RRPRConfoundStrictN{} strict evaluations. For LLM-driven NAS, scaled reporting should include the full generation ledger: successes, failures, target alignment, architecture statistics, confound audits, and sampling budget.

\noindent\textbf{Acknowledgments.} This work was partially supported by the Alexander von Humboldt Foundation.

\bibliographystyle{splncs04}
{\small
\bibliography{references}

@InProceedings{uzun2026closedloop,
  title     = {Closed-Loop {LLM} Discovery of Non-Standard Channel Priors in Vision Models},
  author    = {Uzun, Tolgay Atinc and Ignatov, Dmitry and Timofte, Radu},
  booktitle = {Proceedings of the International Conference on Pattern Recognition (ICPR)},
  year      = {2026},
  note      = {to appear},
  eprint    = {2601.08517},
  archivePrefix = {arXiv},
  url       = {https://arxiv.org/abs/2601.08517}
}

@article{zoph2016neural,
  author  = {Zoph, Barret and Le, Quoc V.},
  title   = {Neural Architecture Search with Reinforcement Learning},
  journal = {arXiv preprint arXiv:1611.01578},
  year    = {2016}
}

@article{liu2018darts,
  author  = {Liu, Hanxiao and Simonyan, Karen and Yang, Yiming},
  title   = {{DARTS}: Differentiable Architecture Search},
  journal = {arXiv preprint arXiv:1806.09055},
  year    = {2018}
}

@article{wang2020channelnet,
  author  = {Yu, Jiahui and Huang, Thomas S.},
  title   = {{AutoSlim}: Towards One-Shot Architecture Search for Channel Numbers},
  journal = {arXiv preprint arXiv:1903.11728},
  year    = {2019}
}

@article{wang2023llmnas,
  author  = {Zheng, Mingkai and Su, Xiu and You, Shan and Wang, Fei and Qian, Chen and Xu, Chang and Albanie, Samuel},
  title   = {Can {GPT-4} Perform Neural Architecture Search?},
  journal = {arXiv preprint arXiv:2304.10970},
  year    = {2023}
}

@article{hu2021lora,
  author  = {Hu, Edward J. and Shen, Yelong and Wallis, Phillip and Allen-Zhu, Zeyuan and Li, Yuanzhi and Wang, Shean and Wang, Lu and Chen, Weizhu},
  title   = {{LoRA}: Low-Rank Adaptation of Large Language Models},
  journal = {arXiv preprint arXiv:2106.09685},
  year    = {2021}
}

@inproceedings{dettmers2023qlora,
  author    = {Dettmers, Tim and Pagnoni, Artidoro and Holtzman, Ari and Zettlemoyer, Luke},
  title     = {{QLoRA}: Efficient Finetuning of Quantized {LLMs}},
  booktitle = {Advances in Neural Information Processing Systems},
  year      = {2023}
}

@inproceedings{xu2023evoprompting,
  author    = {Chen, Angelica and Dohan, David M. and So, David R.},
  title     = {{EvoPrompting}: Language Models for Code-Level Neural Architecture Search},
  booktitle = {Advances in Neural Information Processing Systems},
  year      = {2023}
}

@article{nasir2023llmatic,
  author  = {Nasir, Muhammad U. and Earle, Sam and Cleghorn, Christopher W. and James, Steven and Togelius, Julian},
  title   = {{LLMatic}: Neural Architecture Search via Large Language Models and Quality Diversity Optimization},
  journal = {arXiv preprint arXiv:2306.01102},
  year    = {2023}
}

@article{zhou2024lapt,
  author  = {Zhou, Xun and Wu, Xingyu and Feng, Liang and Lu, Zhichao and Tan, Kay Chen},
  title   = {Design Principle Transfer in Neural Architecture Search via Large Language Models},
  journal = {arXiv preprint arXiv:2408.11330},
  year    = {2024}
}

@article{cai2025seki,
  author  = {Cai, Zicheng and Tang, Yaohua and Lai, Yutao and Wang, Hua and Chen, Zhi and Chen, Hao},
  title   = {{SEKI}: Self-Evolution and Knowledge Inspiration based Neural Architecture Search via Large Language Models},
  journal = {arXiv preprint arXiv:2502.20422},
  year    = {2025}
}

@article{li2025collmnas,
  author  = {Li, Zhe and Lin, Zhiwei and Wang, Yongtao},
  title   = {{CoLLM-NAS}: Collaborative Large Language Models for Efficient Knowledge-Guided Neural Architecture Search},
  journal = {arXiv preprint arXiv:2509.26037},
  year    = {2025}
}

@article{zhang2023gptnas,
  author  = {Yu, Caiyang and Liu, Xianggen and Wang, Yifan and Liu, Yun and Feng, Wentao and Xiong, Deng and Tang, Chenwei and Lv, Jiancheng},
  title   = {{GPT-NAS}: Evolutionary Neural Architecture Search with the Generative Pre-Trained Model},
  journal = {arXiv preprint arXiv:2305.05351},
  year    = {2023}
}

@InProceedings{kochnev2025nngpt,
  title     = {{NNGPT}: Rethinking {AutoML} with Large Language Models},
  author    = {Kochnev, Roman and Khalid, Waleed and Uzun, Tolgay Atinc and Zhang, Xi and Dhameliya, Yashkumar Sanjaybhai and Qin, Furui and Vysyaraju, Chandini and Duvvuri, Raghuvir and Goyal, Avi and Ignatov, Dmitry and Timofte, Radu},
  booktitle = {Proceedings of the IEEE/CVF Conference on Computer Vision and Pattern Recognition (CVPR) Workshops},
  month     = {June},
  year      = {2026},
  pages     = {3262--3272},
  eprint    = {2511.20333},
  archivePrefix = {arXiv},
  url       = {https://openaccess.thecvf.com/content/CVPR2026W/CVPR-NAS26/html/Kochnev_NNGPT_Rethinking_AutoML_with_Large_Language_Models_CVPRW_2026_paper.html}
}

@inproceedings{he2018amc,
  author    = {He, Yihui and Lin, Ji and Liu, Zhijian and Wang, Hanrui and Li, Li-Jia and Han, Song},
  title     = {{AMC}: {AutoML} for Model Compression and Acceleration on Mobile Devices},
  booktitle = {Proceedings of the European Conference on Computer Vision (ECCV)},
  year      = {2018}
}

@inproceedings{liu2017slimming,
  author    = {Liu, Zhuang and Li, Jianguo and Shen, Zhiqiang and Huang, Gao and Yan, Shoumeng and Zhang, Changshui},
  title     = {Learning Efficient Convolutional Networks through Network Slimming},
  booktitle = {Proceedings of the IEEE International Conference on Computer Vision},
  year      = {2017}
}

@inproceedings{liu2019metapruning,
  author    = {Liu, Zechun and Mu, Haoyuan and Zhang, Xiangyu and Guo, Zichao and Yang, Xin and Cheng, Kwang-Ting and Sun, Jian},
  title     = {{MetaPruning}: Meta Learning for Automatic Neural Network Channel Pruning},
  booktitle = {Proceedings of the IEEE/CVF International Conference on Computer Vision},
  year      = {2019}
}

@article{lemur2025,
  author  = {Goodarzi, Arash Torabi and Kochnev, Roman and Khalid, Waleed and Goudarzi, Hojjat Torabi and Qin, Furui and Uzun, Tolgay Atinc and Dhameliya, Yashkumar Sanjaybhai and Kathiriya, Yash Kanubhai and Bentyn, Zofia Antonina and Ignatov, Dmitry and Timofte, Radu},
  title   = {{LEMUR} Neural Network Dataset: Towards Seamless {AutoML}},
  journal = {arXiv preprint arXiv:2504.10552},
  year    = {2025},
  eprint  = {2504.10552},
  archivePrefix = {arXiv},
  url     = {https://arxiv.org/abs/2504.10552}
}

@InProceedings{lemur2_2026,
  title     = {{LEMUR} 2: Unlocking Neural Network Diversity for {AI}},
  author    = {Uzun, Tolgay Atinc and Khalid, Waleed and Din, Saif U and Mulukuledu, Sai Revanth and Singh, Akashdeep and Vysyaraju, Chandini and Duvvuri, Raghuvir and Goyal, Avi and Lukhi, Yashkumar R. and A Hussain, Muhammad and Jesani, Krunal and Shrestha, Usha and Mittal, Yash and Kochnev, Roman and Kadam, Pritam and Ikram, Mohsin and Moradiya, Harsh and Arslanian, Alice and Ignatov, Dmitry and Timofte, Radu},
  booktitle = {Proceedings of the IEEE/CVF Conference on Computer Vision and Pattern Recognition (CVPR) Workshops},
  month     = {June},
  year      = {2026},
  pages     = {3291--3300},
  url       = {https://openaccess.thecvf.com/content/CVPR2026W/CVPR-NAS26/html/Uzun_LEMUR_2_Unlocking_Neural_Network_Diversity_for_AI_CVPRW_2026_paper.html}
}

@article{chee2018airnet,
  title={Airnet: Self-supervised affine registration for 3d medical images using neural networks},
  author={Chee, Evelyn and Wu, Zhenzhou},
  journal={arXiv preprint arXiv:1810.02583},
  year={2018}
}

@misc{penedo2025olympiccoder, title={OlympicCoder}, author={Guilherme Penedo and Anton Lozhkov and Hynek Kydlíček and Loubna Ben Allal and Edward Beeching and Agustín Piqueres Lajarín and Quentin Gallouédec and Nathan Habib and Lewis Tunstall and Leandro von Werra}, year={2025}, publisher = {Hugging Face}, journal = {Hugging Face repository}, howpublished = {\url{https://huggingface.co/open-r1/OlympicCoder-7B}} }

@article{zhou2020econas,
  author  = {Zhou, Dongzhan and Zhou, Xinchi and Zhang, Wenwei and Loy, Chen Change and Yi, Shuai and Zhang, Xuesen and Ouyang, Wanli},
  title   = {{EcoNAS}: Finding Proxies for Economical Neural Architecture Search},
  journal = {arXiv preprint arXiv:2001.01233},
  year    = {2020}
}

@article{abdelfattah2021zerocost,
  author  = {Abdelfattah, Mohamed S. and Mehrotra, Abhinav and Dudziak, {\L}ukasz and Lane, Nicholas D.},
  title   = {Zero-Cost Proxies for Lightweight {NAS}},
  journal = {arXiv preprint arXiv:2101.08134},
  year    = {2021}
}

@article{yu2020evaluating,
  author  = {Yu, Kaicheng and Sciuto, Christian and Jaggi, Martin and Musat, Claudiu and Salzmann, Mathieu},
  title   = {Evaluating the Search Phase of Neural Architecture Search},
  journal = {arXiv preprint arXiv:1902.08142},
  year    = {2019}
}
}

\end{document}